\documentclass[11pt]{article}

\usepackage[final]{acl}

\usepackage{times}
\usepackage{latexsym}
\usepackage{lipsum}
\usepackage{booktabs}
\usepackage[T1]{fontenc}
\usepackage{enumitem}
\usepackage{amsfonts}
\usepackage{amssymb}
\usepackage{todonotes}
\usepackage{fontawesome5} 
\usepackage{xcolor} 
\usepackage{tcolorbox} 

\usepackage[utf8]{inputenc}

\usepackage{microtype}

\usepackage{inconsolata}

\usepackage{graphicx}

\title{DiPS: Dialogue Policy Selection for High-Stakes Persuasion Agents}

\author{Tianyi Zhang$^*$, Mousumi Das\thanks{indicates equal contribution}, \\ 
\textbf{Abrar Anwar, Jesse Thomason, David Traum} \\
  University of Southern California \\
  \texttt{\{tzhang62, mousumid, abrar.anwar, jessetho, traum\}@usc.edu}
  }

\begin{document}
\maketitle
\begin{abstract}
Large Language Models (LLMs) often struggle with persuasion in high-stakes scenarios.
People's individual personalities and concerns require tailored strategies rather than a one-size-fits-all approach. 
To address this challenge, we focus on a fire-rescue scenario in which an operator must persuade a resident to evacuate as a high-stakes persuasion domain and propose \textbf{Di}alogue \textbf{P}olicy \textbf{S}election (DiPS), a Q-learning framework to dynamically select persuasion strategies adapted to the evolving conversational context. 
Specifically, we train a critic, trained to maximize the chance of evacuation success, to select a persuasion policy at each turn based on the resident's recent utterances.
We then evaluate DiPS against multiple baselines in both simulated and real human interactions.
We find that DiPS achieves higher evacuation success than a zero-shot LLM and generic RAG-augmented approach.
\end{abstract}

\definecolor{opcolor}{HTML}{EBF4F5}
\definecolor{rescolor}{HTML}{B5C6E0}

\newtcolorbox{opbox}{
  colback=opcolor, colframe=opcolor,
  arc=8pt, boxrule=0pt,
  left=6pt, right=6pt, top=4pt, bottom=4pt,
  width=0.82\linewidth
}
\newtcolorbox{resbox}{
  colback=rescolor, colframe=rescolor,
  arc=8pt, boxrule=0pt,
  left=6pt, right=6pt, top=4pt, bottom=4pt,
  width=0.82\linewidth
}











\section{Introduction}

\begin{figure}[t]
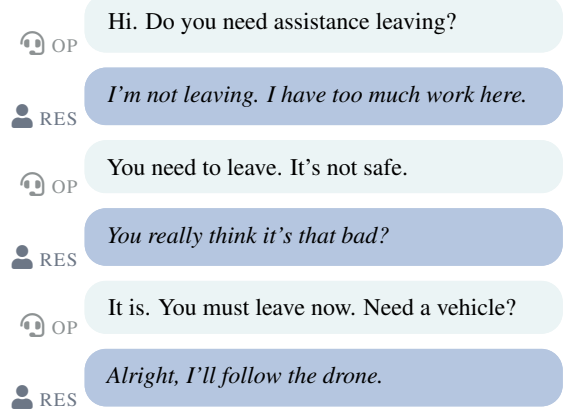

\centering
\small

\setlength{\tabcolsep}{2pt}
\renewcommand{\arraystretch}{0.9}

\begin{tabular}{@{}r@{\;}l@{}}

\textcolor{opcolor!60!black}{\faHeadset\;\textsc{op}} &
\begin{opbox}
Hi. Do you need assistance leaving?
\end{opbox} \\[3pt]

\textcolor{rescolor!60!black}{\faUser\;\textsc{res}} &
\begin{resbox}
\textit{I'm not leaving. I have too much work here.}
\end{resbox} \\[3pt]

\textcolor{opcolor!60!black}{\faHeadset\;\textsc{op}} &
\begin{opbox}
You need to leave. It's not safe.
\end{opbox} \\[3pt]

\textcolor{rescolor!60!black}{\faUser\;\textsc{res}} &
\begin{resbox}
\textit{You really think it's that bad?}
\end{resbox} \\[3pt]

\textcolor{opcolor!60!black}{\faHeadset\;\textsc{op}} &
\begin{opbox}
It is. You must leave now. Need a vehicle?
\end{opbox} \\[3pt]

\textcolor{rescolor!60!black}{\faUser\;\textsc{res}} &
\begin{resbox}
\textit{Alright, I'll follow the drone.}
\end{resbox} \\

\end{tabular}

\caption{Excerpt from a wildfire evacuation dialogue. An operator persuades a reluctant resident to evacuate through escalating urgency and assistance.}
\label{fig:teaser}
\end{figure}

Natural disasters are inherently high-stakes scenarios where decisions about what to do (e.g., whether to evacuate or remain and try to protect one's property) can have severe consequences, up to and including loss of life. 
Disaster relief operators often serve as the primary point of contact with residents in an affected area.
Such operators warn residents of potential problems, giving advice and information, and sometimes providing additional resources, such as transportation, directions, and evacuation facilities (Figure~\ref{fig:teaser}).

Often, there are not enough human operators to effectively communicate with all affected residents. 
Existing work has explored how artificial agents can engage in dialogue with residents in these settings to improve evacuation outcomes~\cite{vaidyanath2020using}. 
However, this kind of high-stakes persuasion is very challenging, since residents will differ (sometimes drastically) in their circumstances, such as how willing and well equipped they might be to evacuate, what might convince them to do so, and what resources they need to do so. 
AI Agents need to be able to adapt from one-size-fits-all protocols to be able to address the specific needs and concerns of individual residents.

We examine several methods for instantiating such an artificial operator. 
These methods vary in terms of what kinds of domain-specific training data they use and how they generate language responses to residents.
We test agent operator methods on both seen and unseen resident scenarios, both in simulation (using an LLM-based resident simulator) and with human participants playing the roles of residents.


In Section~\ref{sec:background} we describe prior work that we build upon, including the simulation environment and dataset we use, as well as other work on adaptive policy creation. We describe the main problem formulation
and our proposed approach in Section~\ref{section:adaptive}. We then introduce three experiments in Section~\ref{sec:experiment_design}, comparing this approach against baselines in dialogue with AI dialogue partners (Section~\ref{sec:exp1}), human roleplayers (Section~\ref{sec:exp2}), and with improved policies for both sides (Section~\ref{sec:exp3}). We conclude in Section~\ref{sec:discussion} with some prospects for future work.

\section{Background and Related Work}
\label{sec:background}
 \subsection{Wildfire Evacuation Persuasion}
\label{sec:wildfire-scenario}
As a testbed for our experiments, we adapt the wildfire simulation setting used in ~\cite{chaffey2020human}. In this simulation, a forest fire is rapidly moving toward a small town. A control center operator must manage several resources including flying drones, a ground vehicle, and an AI assistant to track the fire, locate residents of the town, and convince and help them evacuate. In the original simulation, there were five different resident groups placed randomly within the town. The residents differed in terms of details of their specific identity and backstory, and also their current situation and concerns about whether to evacuate, and what they would need help with to evacuate. For example, some were groups who would want to leave together, some needed a vehicle to help them evacuate, while others needed directions or guidance. Some were willing to evacuate once informed of the danger, while other "stubborn" residents resisted initial attempts and needed empathy and/or convincing of the seriousness of this fire compared to previous near-misses.

\citet{chaffey2020human} also describe a data collection exercise
in which participants in the experiments of humans play the role of the operator, while town residents were controlled by experimenters, using a Wizard of Oz interface to select recorded lines for the resident. The individual dialogues between operator and residents were extracted, transcribed, and annotated~\cite{nasihati2023multimodal,to-etal-2023-common}. 

\subsection{Persuasion algorithms}
Offline reinforcement learning aims to learn effective policies from static, previously collected datasets without requiring further environment interaction \citep{levine2020offline}. A fundamental challenge in this paradigm is distributional shift: the learned policy may overestimate Q-values for out-of-distribution actions not represented in the data \citep{kumar2020conservative, fujimoto2019off}. Implicit Q-Learning (IQL) \citep{kostrikov2021offline} addresses this by avoiding queries over unseen actions entirely, instead using expectile regression to approximate the maximum over actions through a state-value function trained only on in-distribution data. The resulting policy is extracted via advantage-weighted regression, making IQL particularly well-suited for settings with complex action spaces and limited data coverage.

Reinforcement learning has a long history in dialogue systems, from early work on optimizing slot-filling dialogue managers \citep{williams2007partially} to early neural approaches for improving engagement \citep{li2016deep} and task completion \citep{peng2017composite}. While Reinforcement Learning from Human Feedback (RLHF) has become the dominant paradigm for aligning large language models with human preferences \citep{ouyang2022training, bai2022training}, it typically operates at the response level rather than over multi-turn strategic planning. Closer to our work, \citet{verma2022chai} applied offline RL for conversational health agents, and \citet{snell2022offline} explored offline methods for improving multi-turn dialogue outcomes, demonstrating that Q-learning from logged interactions can outperform behavior cloning. Our work shares this motivation but focuses on selecting \emph{which} high-level persuasion strategy to deploy at each turn rather than directly generating utterances.

Recent work has also demonstrated LLMs' capacity for role-playing, where models adopt specific personas or behavioral profiles \citep{shanahan2023role}. \citet{park2023generative} introduced generative agents that simulate believable human behavior, and role-playing has been used to create realistic user simulators for evaluating conversational systems \citep{davidson2023user}. We leverage this capability in evaluation, using persona-conditioned LLMs to simulate residents in automated experiments and human participants who adopt resident personas in a live study.

In the domain of computational persuasion, prior work spans health behavior change \citep{wang2019persuasion}, negotiation \citep{he2018decoupling, lewis2017deal}, and debate \citep{durmus2019role}. \citet{wang2019persuasion} introduced PersuasionForGood and explored strategy prediction for charitable donation dialogues, while \citet{salvi2024conversational} examined the persuasive capabilities of GPT-4 across multiple domains. However, most existing approaches either apply a single persuasion strategy uniformly or rely on LLMs' implicit persuasive abilities without explicit strategic planning. In contrast, our framework, DiPS, learns to \emph{dynamically} select among distinct persuasion policies based on the evolving conversational context. By applying IQL to fire-rescue evacuation persuasion, where the stakes are life-threatening, we demonstrate that offline RL can effectively learn turn-level policy selection from heterogeneous dialogue data. To our knowledge, this represents the first application of implicit Q-learning to high-stakes persuasive dialogue, bridging offline RL methods and the nuanced demands of adaptive human persuasion.

\section{Dialogue Policy Selection} \label{section:adaptive}
We study \emph{adaptive persuasion} in high-stakes dialogue, where an operator must dynamically tailor responses to an individual’s evolving behavior under partial observability. 
Consider the emergency evacuation scenario described in Section~\ref{sec:wildfire-scenario}. Even when the risk is clear, individuals respond differently: some may be skeptical, others emotionally attached to their homes, and others overwhelmed or indecisive. Uniform instructions (e.g., repeatedly asking a resident to evacuate) often fail to account for these differences, leading to disengagement or resistance~\cite{nasihati2023multimodal}. Effective persuasion in such settings requires adapting both the content and tone of responses to the resident’s current state, rather than relying on a fixed strategy.

\subsection{Notation and Problem Definition}

\paragraph{Dialogue Model.}
We model the interaction as a partially observable Markov decision process (POMDP):
\[
\mathcal{M} = (\mathcal{S}, \mathcal{O}, \mathcal{A}, \mathcal{P}, \mathcal{R}, \gamma).
\]
At each timestep $t$, the resident is in a latent state $s_t \in \mathcal{S}$ capturing internal factors such as beliefs, emotional state, and risk perception. 
The operator observes $o_t \in \mathcal{O}$, representing the resident’s response utterance to previous operator utterance $a_{t-1}$.
The operator additionally maintains a dialogue history $h_t = (a_0,o_1, a_1, \dots, o_t)$. 
Given $h_t$, the operator policy selects a response $\pi(h_t)=a_t \in \mathcal{A}$, after which the state evolves as ${s_{t+1} \sim \mathcal{P}(\cdot \mid s_t, a_t)}$.

The objective is to maximize expected discounted ($\gamma$) future return, given by reward function $R:S\rightarrow\mathbb{R}$, over a finite horizon $T$, corresponding to successful task completion:
\[
\max_{\pi} \; \mathbb{E}_{\pi, \mathcal{P}} \left[ \sum_{t=1}^T \gamma^{t-1} r_t \right],
\]
where $r_t$ denotes $R(s_t)$ for ${s_t\sim\mathcal{P}(\cdot\mid s_{t-1},\pi(h_{t-1}))}$ and preceding states starting from initial, empty dialogue state $s_0$.

\paragraph{Adaptive Persuasion via Policy Classes.}
Rather than learning a single monolithic policy, we assume access to a set of $K$ candidate persuasion policies (or personas)
\[
\Pi = \{\pi^{(1)}, \dots, \pi^{(K)}\},
\]
where each $\pi^{(k)}(a_t \mid h_t)$ encodes a distinct strategy (e.g., empathetic, directive, or informational). At each timestep, a policy index $z_t \in \{1, \dots, K\}$ is selected, and the resulting action is drawn as
\[
a_t \sim \pi^{(z_t)}(\cdot \mid h_t).
\]

\paragraph{Problem Statement.}
Given a set of candidate persuasion policies $\Pi$ and a dataset of dialogues, the goal of adaptive persuasion is to learn a policy over policies that selects, at each timestep, the most effective strategy conditioned on the dialogue history.

\begin{figure*}
    \centering
    \includegraphics[width=1\linewidth]{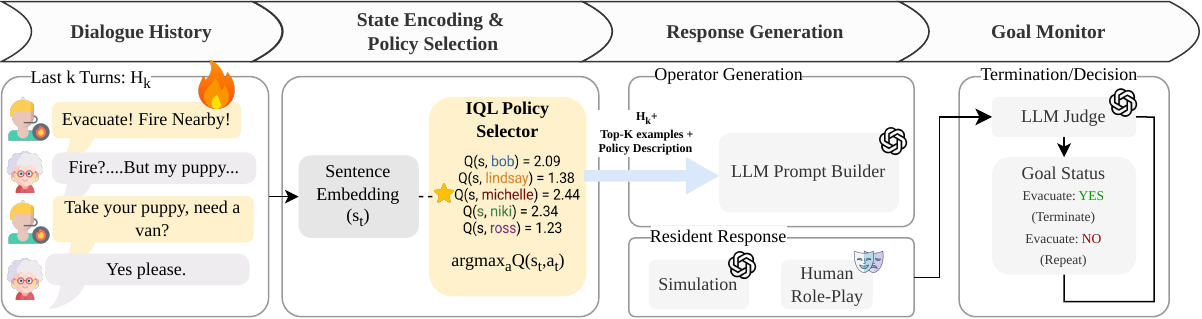}
\caption{\textbf{DiPS for adaptive persuasion}. Given the recent dialogue history, a state encoder produces a representation of the conversation, which is used by an IQL-trained Q-function to select a persuasion policy (persona) at each turn. The selected policy conditions use strategy descriptions and retrieved examples to prompt the LLM to generate an operator response. The resident (simulated or human) replies, and the interaction continues until a termination decision is made by an LLM judge or the (human) resident ends the conversation.}
    \label{fig:main_}
\end{figure*}

\subsection{DiPS Framework.}
\label{sec:DiPS}
We introduce \textbf{Di}alogue \textbf{P}olicy \textbf{S}election (DiPS), a framework for adaptive persuasion that learns to dynamically select among a set of policies. Rather than directly optimizing over natural language responses, DiPS operates at a higher level, where each action corresponds to selecting a persona used to condition the response generation.

\paragraph{Offline Dialogue Dataset.}
We assume access to an offline dataset of dialogues $\mathcal{D}$ collected across multiple personas. Each trajectory $\tau \in \mathcal{D}$ consists of a sequence
$\tau = (h_1, z_1, h_2, z_2, \dots, h_T),$
where $h_t$ is the dialogue history and $z_t \in \{1, \dots, K\}$ denotes the persona used at time $t$. The dataset contains both successful and unsuccessful interactions, providing signal about which strategies are effective under different conversational contexts. Each transition yields tuples $(h_t, z_t, r_t, h_{t+1})$, where rewards $r_t$ reflect downstream task success.

\paragraph{Policy Selection via Offline RL.}
We formulate dialogue policy selection as an offline reinforcement learning problem over the discrete action space of personas. The goal is to learn a value function $Q(h_t, z)$ that estimates the long-term effectiveness of selecting persona $z$ given the current dialogue history.

We train $Q$ using Implicit Q-Learning (IQL), which enables stable learning from offline data without requiring additional interaction. IQL learns to assign higher values to personas that lead to successful outcomes while down-weighting those observed in failed trajectories. Intuitively, this learning leads the model to infer which strategies are effective for different types of residents and conversational contexts.

\paragraph{Policy Extraction.}
At inference time, DiPS selects the persona $z_t$ at a given timestep based on the Q-function:
\[
z_t = \arg\max_{z \in \{1, \dots, K\}} Q(h_t, z).
\]
The selected persona is used to generate a response via the corresponding policy $\pi^{(z_t)}(a_t \mid h_t)$.

Operating over a discrete set of strategies and leveraging offline reinforcement learning, DiPS learns when each persuasion style should be deployed. Persona switching enables dynamic adaptation throughout the dialogue; the system can shift tone and strategy in response to the resident’s behavior, rather than relying on a fixed policy.

\subsection{Application to Simulated Fire Rescue}
To apply DiPS to the 
scenario described in Section~\ref{sec:wildfire-scenario}, we adapted existing data as follows. A summary of the pipeline for DiPS applied to this domain is shown in Figure~\ref{fig:main_}.

\paragraph{Residents and Operator Policies.}
We consider a set of residents $\{R_1, \dots, R_N\}$, each corresponding to a distinct behavioral profile (e.g., skeptical, emotional, or resistant). We additionally construct a set of operator policies $\Pi = \{\pi_1, \dots, \pi_K\}$, where each policy represents a consistent persuasion strategy derived from a specific operator’s interaction style (e.g., empathetic, directive, or assertive).

Each $\pi_k$ is instantiated from dialogue data generated by a human or simulated operator interacting with residents. Importantly, these policies are reused across residents, enabling cross-resident generalization.
In particular, we define $\pi_k$ as a persona-specific addition to an LLM-driven operator utterance generator, as well as RAG in-context learning examples of dialogue histories and operator responses in the Wizard-of-Oz dialogues involving resident $k$.

\paragraph{Dataset Construction.}
We construct an offline reinforcement learning dataset from evacuation dialogues by converting each conversation into transition tuples $(s_t, a_t, r_t, s_{t+1})$. The state $s_t$ is defined as a sentence embedding of recent resident utterances (using MiniLM), and the action $a_t$ corresponds to a discrete operator policy.
Note that the terminology for states and rewards is equivalent to the operator POMDP formulation detailed above; here, the action $a_t$ corresponds not to an operator message to the resident, but to a latent persona selection used to condition an LLM-driven operator agent to generate that message. 

Rewards are sparse: a value of $1$ is assigned at the final turn of a successful dialogue (i.e., when the resident agrees to evacuate), and $0$ otherwise. To evaluate generalization, we partition residents into disjoint training and test sets. The policy selector is trained only on training residents and evaluated on unseen residents with distinct behavioral profiles.
With this offline data formulation, note that each training dialogue corresponds to taking a single action at each timestep---that action corresponds to the operator persona being learned to effectively communicate with the specific resident the dialogue is with.

\section{Experiment Design}
\label{sec:experiment_design}
We conducted three sets of experiments to explore the ability of AI agents to play the role of an operator in wildfire persuasion dialogues, and more specifically to test DiPS in this environment against non-adaptive policies. For each experiment, we use the same test-bed: text-based dialogues in which a policy-driven operator interacts with and tries to convince a resident to evacuate. We use ten resident profiles: five drawn from prior work~\cite{chaffey2020human,nasihati2023multimodal} and five newly constructed profiles with similar concerns but distinct backstories. Each interaction is a multi-turn dialogue, and success is defined by whether the resident ultimately agrees to evacuate. 

In the first experiment, described in Section~\ref{sec:exp1}, we compare a first version of DiPS with two other policies, \textit{zero-shot}, which does not make use of prior data, and {\em RAG}, which makes use of the same data as DiPS, but does not partition it into individual personas and is free to draw from any utterances. This experiment involves simulated residents, using information similar to what the Wizards used in playing the residents in~\cite{chaffey2020human}. In the second experiment, described in Section~\ref{sec:exp2}, we evaluated the same three policies with human roleplayers as residents. On the basis of findings in these experiments, we conducted a third experiment, described in Section~\ref{sec:exp3}, with an expanded set of operator policies, including some using additional types of information, as well as improving the resident simulators.

\section{Experiment 1: AI Roleplay Residents}
\label{sec:exp1}


\paragraph{Simulation Framework.}
We conduct experiments in a controlled simulation environment with two LLM-based simulators: a \emph{resident simulator} and an \emph{operator simulator}. The resident simulator is conditioned on structured persona descriptions and generates responses following a staged disclosure pattern, where key concerns are revealed progressively.
The operator simulator generates responses turn-by-turn using a shared dialogue engine across all methods. 



\paragraph{Simulated Dialogues.}
We simulate resident--operator interactions using LLM-based agents. GPT-4o-mini is used as the backbone model for both the resident and operator simulators. Each resident is conditioned on a structured persona and dialogue history. For each combination of method (Zero-shot, RAG, DiPS) and resident, multiple dialogue runs are collected, yielding 90 simulated conversations in total. Dialogues are capped at a maximum of 15 turns, and both operator and resident models use a temperature of 0.2. Additional details on prompts and resident profiles are provided in Appendix~\ref{sec:resident_profiles} and~\ref{sec:prompts}.

\paragraph{Simulated Dialogue Evaluation.}
Each simulated dialogue is evaluated using an LLM-as-judge (GPT-4o-mini), which determines whether the resident successfully evacuates based on the full transcript. This binary outcome defines the terminal reward. 
To validate this evaluation, a subset of dialogues is additionally labeled by two human annotators using a three-way scheme: ``yes'' (explicit agreement with a viable plan), ``maybe'' (uncertain or conditional outcome), and ``no'' (explicit refusal or infeasible conditions). The annotators achieve over 80\% agreement. The same labeling instructions are used for both human annotators and the LLM judge (see Appendix~\ref{sec:labeling}). For each condition, three dialogues per resident (90 total) are evaluated by both human annotators and the LLM.

\begin{table*}[t]
\centering
\setlength{\tabcolsep}{4pt}
\small
\begin{tabular}{lrrrr|rrrr}
\toprule
 & \multicolumn{4}{c|}{\textbf{Simulation}} & \multicolumn{4}{c}{\textbf{Human Subject}} \\
 & Zero-shot & RAG & DiPS & All & Zero-shot & RAG & DiPS & All \\
\midrule
\multicolumn{9}{l}{\textit{Existing residents in WoZ data}} \\
Bob & 3/3 (100\%) & 3/3 (100\%) & 1/3 (\phantom{1}33\%) & 7/9 (78\%) & 1/3 (33\%) & 3/3 (100\%) & 3/3 (100\%) & 7/9 (78\%) \\
Lind. & 2/3 (\phantom{1}67\%) & 2/3 (\phantom{1}67\%) & 3/3 (100\%) & 7/9 (78\%) & 13/15 (87\%) & 10/15 (\phantom{1}67\%) & 14/15 (\phantom{1}93\%) & 37/45 (82\%) \\
Niki & 3/3 (100\%) & 2/3 (\phantom{1}67\%) & 2/3 (\phantom{1}67\%) & 7/9 (78\%) & 1/2 (50\%) & 1/2 (\phantom{1}50\%) & 1/2 (\phantom{1}50\%) & 3/6 (50\%) \\
Ross & 1/3 (\phantom{1}33\%) & 0/3 (\phantom{10}0\%) & 1/3 (\phantom{1}33\%) & 2/9 (22\%) & 3/6 (50\%) & 4/6 (\phantom{1}67\%) & 2/6 (\phantom{1}33\%) & 9/18 (50\%) \\
\cmidrule{2-9}
\textit{Avg} & 75\% & 58\% & 58\% & 64\% & 55\% & 71\% & 69\% & 65\% \\
\midrule
\multicolumn{9}{l}{\textit{New residents unseen by policy conditioning}} \\
Ana & 1/3 (\phantom{1}33\%) & 3/3 (100\%) & 2/3 (\phantom{1}67\%) & 6/9 (\phantom{1}67\%) & 5/6 (\phantom{1}83\%) & 4/6 (\phantom{1}67\%) & 6/6 (100\%) & 15/18 (\phantom{1}83\%) \\
Ben & 3/3 (100\%) & 2/3 (\phantom{1}67\%) & 3/3 (100\%) & 8/9 (\phantom{1}89\%) & 7/8 (\phantom{1}88\%) & 7/8 (\phantom{1}88\%) & 6/8 (\phantom{1}75\%) & 20/24 (\phantom{1}83\%) \\
Mary & 3/3 (100\%) & 3/3 (100\%) & 0/3 (\phantom{10}0\%) & 6/9 (\phantom{1}67\%) & 6/8 (\phantom{1}75\%) & 2/8 (\phantom{1}25\%) & 7/8 (\phantom{1}88\%) & 15/24 (\phantom{1}62\%) \\
Mia & 3/3 (100\%) & 3/3 (100\%) & 2/3 (\phantom{1}67\%) & 8/9 (\phantom{1}89\%) & 4/4 (100\%) & 2/4 (\phantom{1}50\%) & 3/4 (\phantom{1}75\%) & 9/12 (\phantom{1}75\%) \\
Tom & 3/3 (100\%) & 3/3 (100\%) & 3/3 (100\%) & 9/9 (100\%) & 2/2 (100\%) & 2/2 (100\%) & 2/2 (100\%) & 6/6 (100\%) \\
\cmidrule{2-9}
\textit{Avg} & 87\% & 93\% & 67\% & 82\% & 89\% & 66\% & 88\% & 81\% \\
\midrule
\textbf{Avg} & 81\% & 78\% & 63\% & 74\% & 74\% & 68\% & 79\% & 74\% \\
\bottomrule
\end{tabular}
\caption{Initial simulation and human subject results. Success rate by character and conversation type. Simulation success is determined by human labeling of dialogue outcomes; human-subject success is measured by whether the participant agreed to evacuate during the dialogue.}

\label{tab:success-combined}
\end{table*}

\subsection{Results}

The left side of Table~\ref{tab:success-combined} shows the results from Experiment 1. Overall success rates are comparable to prior human-operator studies~\cite{nasihati2023multimodal}, which report an 85\% acceptance rate. However, performance varies significantly across residents and methods. 

Surprisingly, zero-shot performs best on the resident profiles from existing WoZ data, despite not having access to training data for them.
RAG performs best on the new residents, suggesting that retrieval-based methods can adapt effectively to unseen profiles. 
By contrast, DiPS underperforms on both sets.

Several factors may explain these differences. First, the resident simulator may not accurately reflect human behavior. Second, the DiPS prompts may not provide sufficient information for effective policy selection. Third, there may be limitations in applying DiPS to this domain; for example, switching personas mid-dialogue could introduce incoherence and reduce persuasiveness.

\subsection{LLM-as-Judge}

To complement human evaluation, we use an LLM-as-judge (GPT-4o-mini) for both policy training and evaluation. Table~\ref{tab:agreement} reports agreement between the LLM judge and expert human annotators. Overall agreement is 72.2\%, with a Cohen’s $\kappa$ of 0.42.

Agreement varies across methods. DiPS achieves the highest agreement (83.3\%, $\kappa=0.67$), indicating that its outcomes are more clearly interpretable. In contrast, Zero-shot and RAG show lower agreement (66.7\%, $\kappa=0.22$ and $0.30$). 
Dialogues produced by the non-personalized baselines contain more borderline cases where the resident's commitment to evacuate is open to interpretation.

The LLM judge is consistently more conservative than human annotators, assigning lower success rates overall (55.6\% vs.\ 67.8\%). 
In many cases, the LLM labels conditional or incomplete agreements as ``maybe,'' where humans label them as successful (see Appendix~\ref{fig:sample_simulated_dial}).

Overall, while the LLM-as-judge provides a scalable evaluation signal, it should not be treated as a direct substitute for human evaluation, particularly for ambiguous interactions.

\begin{table}[t]
\small
\centering
\begin{tabular}{lcccc}
\toprule
 & \multicolumn{2}{c}{\textbf{Agreement}} & \multicolumn{2}{c}{\textbf{Success rate (\%)}} \\
\cmidrule(lr){2-3} \cmidrule(lr){4-5} 
\textbf{Setting} & \textbf{(raw \%)} & \textbf{$\kappa$} & \textbf{Human} & \textbf{LLM} \\
\midrule
Zero-shot        & 66.7 & 0.22 & 76.7 & 63.3 \\
RAG              & 66.7 & 0.30 & 70.0 & 56.7 \\
DiPS              & 83.3 & 0.67 &  56.7 & 46.7 \\
\midrule
Overall          & 72.2 & 0.42 &  67.8 & 55.6 \\
\bottomrule
\end{tabular}
\caption{Agreement between LLM (gpt-4o-mini) judge and human annotator across experimental settings in simulation ($n=30$ conversations per setting). Agreement collapse \textit{no} and \textit{maybe} into \textit{unsuccessful} vs.\ \textit{successful}; }
\label{tab:agreement}
\end{table}

\section{Experiment 2: Human Residents}
\label{sec:exp2}
%
We recruit human participants to play the role of residents in a web-based chat interface.\footnote{Participants were compensated and the study was registered with and approved by our institutions IRB office.} Each participant selects a character and interacts with all three operator methods in randomized order. After each interaction, participants report whether they were convinced to evacuate and rate dialogue coherence on a 1--5 Likert scale. Operator responses are generated using the same prompting setup as in Experiment 1.


\subsection{Results}

We see a similar overall acceptance rate as for the simulated residents, for both old and new resident profiles (Table~\ref{tab:success-combined} right hand side).
However, the distributions across models and across some characters is quite different.
Here, as we expected, zero-shot did much better with unseen residents, while RAG performed better with residents matching those present in the WoZ training data. 
DiPS did well with both sets of residents, leading to an overall best performance.
%
%
%
At the individual level, several characters exhibit strong reversals: Bob and Mary are easy for Zero-shot and RAG in simulation but substantially easier for DiPS with human participants. 

Given that DiPS underperforms in simulation despite its effectiveness with humans, we hypothesized that the simulation environment itself may be masking the benefits of personalization. 
In Experiment 3, we improved our user simulator based on these human evaluation experiments and conducted additional simulation experiments with several DiPS variations, aiming to identify which components of the personalization pipeline contribute most to dialogue persuasion success and whether the simulation gap can be narrowed.


\subsection{Comparison Analysis: Human vs.\ LLM Operator}
We also examine how these dialogues with automated operator compare to the  data from~\cite{nasihati2023multimodal,to-etal-2023-common} with human operators and wizards roleplaying the residents.



Table~\ref{tab:human-vs-llm-char} shows per-character acceptance rates. While outcomes are similar for Bob and Lindsay, large gaps emerge for Niki ($50\%$ vs.\ $100\%$, $p = .044$) and Ross ($50\%$ vs.\ $93\%$, $p = .001$), driving the overall difference ($72\%$ vs.\ $89\%$, $p = .043$). These discrepancies are concentrated in scenarios requiring coordination or concrete commitments. For example, Ross requires logistical follow-through (e.g., dispatching a vehicle), and Niki requires reassurance and coordination. In these cases, LLM operators often produce plausible language but fail to deliver actionable or coherent commitments, leading to breakdowns (see Appendix~\ref{appendix:dialogue-comparison}.) 

A similar pattern appears within LLM-based interactions: success remains high when participants follow expected scripts ($77$--$84\%$), but drops significantly when they introduce new demands or deviate from expected behavior ($47$--$60\%$). 

These results highlight two consistent gaps: (i) a \emph{logistical commitment gap}, where LLMs fail to follow through on concrete actions, and (ii) an \emph{off-script adaptation gap}, where performance degrades under unexpected interaction patterns. Rather than indicating uniformly weaker performance, these findings identify specific capabilities that human operators handle implicitly. 
These findings motivate our iterated simulation design in Experiment 3, which explicitly incorporates such constraints to better align simulation with human behavior.

\begin{table}[t]
\centering
\small
\setlength{\tabcolsep}{3pt}
\begin{tabular}{l cr cr l}
\toprule
 & \multicolumn{2}{c}{\textbf{Human Operator}}
 & \multicolumn{2}{c}{\textbf{LLM Operator}}
 & \\
\cmidrule(lr){2-3}\cmidrule(lr){4-5}
  & Convinced/\textit{N} & Rate
  & Convinced/\textit{N} & Rate
  & \textit{p} \\
\midrule
Bob     & 20/26 & 77\% & 7/9   & 78\% & 1.000 \\
Lind. & 12/12 & 100\% & 37/45 & 82\% & 0.183 \\
Niki    & 9/9   & 100\% & 3/6   & 50\% & 0.044$^{*}$ \\
Ross    & 27/29 & 93\%  & 9/18  & 50\% & 0.0001$^{**}$ \\
\midrule
& \textbf{68/76} & \textbf{89\%}
                 & \textbf{56/78} & \textbf{72\%}
                 & \textbf{.043}$^{*}$ \\
\bottomrule
\end{tabular}
\caption{Agreement rates by character for the human operator and the LLM operator (pooled across zero-shot, RAG, and DiPS settings). Fisher's exact test assesses whether the two proportions differ significantly for each character. $^{*}p<.05$;\; $^{**}p<.01$.}
\label{tab:human-vs-llm-char}
\end{table}


\section{Experiment 3: Improved Simulations}
\label{sec:exp3}

Motivated by the discrepancies observed in earlier experiments, we refine several components of the simulation environment:

\textbf{Resident simulator.} We augment the resident prompts with explicit \emph{barrier information} that encodes each character’s priorities and constraints. For example, Ross now requires a concrete transportation plan before agreeing to evacuate, while Bob requires an explicit comparison between the risk of staying and the cost of leaving work. This explicit encoding of resident needs in the simulator prompt encourages more consistent and realistic resistance to evacuation.
We also improve the LLM-as-judge setup by simplifying the instructions and shortening the prompt, then manually verifying the decision results.
We found the improved judge had roughly a 90\% agreement with the actual decision made by a human judge.


\textbf{DiPS operator models.} We enhance the operator prompts with a \emph{policy description} that provides richer context about the selected strategy, including structured information about resident concerns. This information is derived from training data and improves performance when aligned with the correct character, though it may introduce risk when applied to mismatched profiles.
Experimentally, we find that these descriptions are valuable even when applied to dialogues with unseen residents. 


\textbf{Ablations.} We include ablations of components of both baseline models and DiPS to better isolate the contributions of policy selection, retrieval, and persona conditioning under the improved simulation.

\subsection{Baselines}

We compare DiPS against a range of non-adaptive and partially adaptive baselines to isolate the effect of learned policy selection.
These include the same \textit{Zero-shot} and \textit{RAG (Global Retrieval)} baselines as the previous experiments, but also the following:


\paragraph{Random Policy Selection.}
We consider several variants where a persona is sampled uniformly at random at each turn. These include:
(i) \emph{Random (All Personas)}, which samples from the full set $\Pi$,
(ii) \emph{Random (Subset Personas)}, which samples from a smaller subset, and
(iii) variants that combine random selection with retrieval (e.g., Random + RAG).
These baselines isolate the contribution of policy selection versus naive selection.

\paragraph{Mismatched Policy.}
A fixed policy is used that does not match the resident’s behavioral profile. This baseline highlights the impact of applying an inappropriate strategy throughout the dialogue.

\subsection{Results}


Table~\ref{tab:main_results} reports results under the improved simulation. Under this setting, DiPS achieves the highest overall success rate (92\%), outperforming Zero-shot (88\%) and RAG (78\%). This trend now aligns with the human study, indicating that adaptive policy selection is more effective when evaluated under a more realistic environment.
In addition to higher success rates, DiPS also requires fewer dialogue turns (11.2 on average), indicating more efficient persuasion. 
These results suggest that evaluation environments that do not model resident persistent resistance or strategy sensitivity can obscure the benefits of adaptive methods.

\subsection{Ablation Studies}

Table~\ref{tab:ablations} shows that removing components consistently degrades performance. Replacing per-policy retrieval with global retrieval reduces success to 86\% and increases dialogue length, while removing retrieval entirely further drops performance to 80\%. Similarly, removing explicit strategy prompting also leads to a comparable drop, despite retaining policy selection.

These results indicate that DiPS relies on the combination of policy selection, persona conditioning, and structured retrieval, with each component contributing to both effectiveness and efficiency.

\begin{table}[t]
\centering
\caption{Comparing DiPS against baselines. DiPS achieves the highest success rate while requiring the fewest turns, indicating both more effective and efficient persuasion. Fixed strategies (zero-shot, RAG) and random policy selection perform significantly worse, with the mismatched policy baseline highlighting the cost of selecting inappropriate strategies.}\label{tab:main_results}
\scriptsize
\begin{tabular}{lcc}
\toprule
Experiment & Overall & AvgTurns \\
\midrule
\textbf{DiPS}        & \textbf{92} & \textbf{11.2} \\
\midrule
Zero-shot                  & 88 & 11.9 \\
RAG                        & 78 & 11.8 \\
\midrule
Random (All Personas)      & 78 & 11.8 \\
Random (Subset Personas)   & 72 & 12.3 \\
Random + RAG (Persona)     & 71 & 12.9 \\
Random + RAG (No Persona)  & 69 & 12.5 \\
Mismatched Policy          & 51 & 13.6 \\
\bottomrule
\end{tabular}
\end{table}

\begin{table}[t]
\centering
\caption{Ablations of DiPS components. Removing components degrades both success rate and efficiency, highlighting the importance of policy selection and persona-conditioned strategies.}
\label{tab:ablations}
\scriptsize
\begin{tabular}{lcc}
\toprule
Experiment & Overall & AvgTurns \\
\midrule
\textbf{DiPS (Full)}              & \textbf{92} & \textbf{11.2} \\
DiPS (Global Retrieval)           & 86 & 13.0 \\
DiPS (Persona Only)               & 80 & 12.9 \\
DiPS (No Persona Prompt)         & 80 & 12.9 \\
\bottomrule
\end{tabular}
\end{table}

\section{Discussion and Future Work}
\label{sec:discussion}
Our experiments have shown that an AI operator agent can be used to successfully persuade residents, at a rate similar to human operators (in a disaster simulation). Adaptive policies, and DiPS in general prove the most successful, when given explicit instructions about likely challenges to overcome, even for unseen characters. LLM-as-judge has reasonable correlation with human judgment, but is not consistent across all conditions.

In future work, we hope to re-evaluate the latest models with human residents, both in the text modality used here as well as the full simulation, using spoken language and under time pressure from the (stochastic) progression of the fire, as in~\cite{chaffey2020human}. 

\section*{Limitations}

Our evaluation relies on LLM-based residents and judges, which, while scalable and consistent, may not fully reflect real human behavior in high-stakes settings. DiPS also operates over a fixed set of predefined personas, limiting expressivity to discrete strategies and potentially missing more nuanced responses. Additionally, the method depends on offline dialogue data, and performance may degrade if certain resident behaviors or failure modes are underrepresented. The learning dynamics are affected by rewards primarily derived from terminal outcomes, making fine-grained credit assignment across dialogue turns challenging.
Finally, all experiments were conducted only in English, using an English language WoZ dataset as reference data, limiting the applicability of findings without multilingual verification.

\clearpage
\section{Acknowledgements}
This work was supported by a grant from the Army
Research Lab (ARL) Army AI Innovations Institute (A2I2), award number W911NF-23-2-0010 and Cooperative Agreement Number W911NF-20-2-
0053.

\bibliography{custom}

@inproceedings{levine2020offline,
 author = {Kumar, Aviral and Zhou, Aurick and Tucker, George and Levine, Sergey},
 booktitle = {Advances in Neural Information Processing Systems},
 editor = {H. Larochelle and M. Ranzato and R. Hadsell and M.F. Balcan and H. Lin},
 pages = {1179--1191},
 publisher = {Curran Associates, Inc.},
 title = {Conservative Q-Learning for Offline Reinforcement Learning},
 url = {https://proceedings.neurips.cc/paper_files/paper/2020/file/0d2b2061826a5df3221116a5085a6052-Paper.pdf},
 volume = {33},
 year = {2020}
}

@article{kumar2020conservative,
  title={Conservative q-learning for offline reinforcement learning},
  author={Kumar, Aviral and Zhou, Aurick and Tucker, George and Levine, Sergey},
  journal={Advances in neural information processing systems},
  volume={33},
  pages={1179--1191},
  year={2020}
}

@inproceedings{fujimoto2019off,
  title={Off-policy deep reinforcement learning without exploration},
  author={Fujimoto, Scott and Meger, David and Precup, Doina},
  booktitle={International conference on machine learning},
  pages={2052--2062},
  year={2019},
  organization={PMLR}
}

@article{kostrikov2021offline,
  title={Offline reinforcement learning with implicit q-learning},
  author={Kostrikov, Ilya and Nair, Ashvin and Levine, Sergey},
  journal={arXiv preprint arXiv:2110.06169},
  year={2021}
}

@article{williams2007partially,
  title={Partially observable Markov decision processes for spoken dialog systems},
  author={Williams, Jason D and Young, Steve},
  journal={Computer Speech \& Language},
  volume={21},
  number={2},
  pages={393--422},
  year={2007},
  publisher={Elsevier}
}

@inproceedings{li2016deep,
  title={Deep reinforcement learning for dialogue generation},
  author={Li, Jiwei and Monroe, Will and Ritter, Alan and Jurafsky, Dan and Galley, Michel and Gao, Jianfeng},
  booktitle={Proceedings of the 2016 conference on empirical methods in natural language processing},
  pages={1192--1202},
  year={2016}
}

@inproceedings{peng2017composite,
  title={Composite task-completion dialogue policy learning via hierarchical deep reinforcement learning},
  author={Peng, Baolin and Li, Xiujun and Li, Lihong and Gao, Jianfeng and Celikyilmaz, Asli and Lee, Sungjin and Wong, Kam-Fai},
  booktitle={Proceedings of the 2017 conference on empirical methods in natural language processing},
  pages={2231--2240},
  year={2017}
}

@article{ouyang2022training,
  title={Training language models to follow instructions with human feedback},
  author={Ouyang, Long and Wu, Jeffrey and Jiang, Xu and Almeida, Diogo and Wainwright, Carroll and Mishkin, Pamela and Zhang, Chong and Agarwal, Sandhini and Slama, Katarina and Ray, Alex and others},
  journal={Advances in neural information processing systems},
  volume={35},
  pages={27730--27744},
  year={2022}
}

@article{bai2022training,
  title={Training a helpful and harmless assistant with reinforcement learning from human feedback},
  author={Bai, Yuntao and Jones, Andy and Ndousse, Kamal and Askell, Amanda and Chen, Anna and DasSarma, Nova and Drain, Dawn and Fort, Stanislav and Ganguli, Deep and Henighan, Tom and others},
  journal={arXiv preprint arXiv:2204.05862},
  year={2022}
}

@inproceedings{verma2022chai,
  title={Chai: A chatbot ai for task-oriented dialogue with offline reinforcement learning},
  author={Verma, Siddharth and Fu, Justin and Yang, Sherry and Levine, Sergey},
  booktitle={Proceedings of the 2022 Conference of the North American Chapter of the Association for Computational Linguistics: Human Language Technologies},
  pages={4471--4491},
  year={2022}
}

@article{snell2022offline,
  title={Offline rl for natural language generation with implicit language q learning},
  author={Snell, Charlie and Kostrikov, Ilya and Su, Yi and Yang, Mengjiao and Levine, Sergey},
  journal={arXiv preprint arXiv:2206.11871},
  year={2022}
}

@article{shanahan2023role,
  title={Role play with large language models},
  author={Shanahan, Murray and McDonell, Kyle and Reynolds, Laria},
  journal={Nature},
  volume={623},
  number={7987},
  pages={493--498},
  year={2023},
  publisher={Nature Publishing Group UK London}
}

@inproceedings{park2023generative,
  title={Generative agents: Interactive simulacra of human behavior},
  author={Park, Joon Sung and O'Brien, Joseph and Cai, Carrie Jun and Morris, Meredith Ringel and Liang, Percy and Bernstein, Michael S},
  booktitle={Proceedings of the 36th annual acm symposium on user interface software and technology},
  pages={1--22},
  year={2023}
}

@article{davidson2023user,
  title={User simulation with large language models for evaluating task-oriented dialogue},
  author={Davidson, Sam and Romeo, Salvatore and Shu, Raphael and Gung, James and Gupta, Arshit and Mansour, Saab and Zhang, Yi},
  journal={arXiv preprint arXiv:2309.13233},
  year={2023}
}

@inproceedings{wang2019persuasion,
  title={Persuasion for good: Towards a personalized persuasive dialogue system for social good},
  author={Wang, Xuewei and Shi, Weiyan and Kim, Richard and Oh, Yoojung and Yang, Sijia and Zhang, Jingwen and Yu, Zhou},
  booktitle={Proceedings of the 57th annual meeting of the association for computational linguistics},
  pages={5635--5649},
  year={2019}
}

@inproceedings{lewis2017deal,
  title={Deal or no deal? end-to-end learning of negotiation dialogues},
  author={Lewis, Mike and Yarats, Denis and Dauphin, Yann and Parikh, Devi and Batra, Dhruv},
  booktitle={Proceedings of the 2017 Conference on Empirical Methods in Natural Language Processing},
  pages={2443--2453},
  year={2017}
}

@inproceedings{durmus2019role,
  title={The role of pragmatic and discourse context in determining argument impact},
  author={Durmus, Esin and Ladhak, Faisal and Cardie, Claire},
  booktitle={Proceedings of the 2019 Conference on Empirical Methods in Natural Language Processing and the 9th International Joint Conference on Natural Language Processing (EMNLP-IJCNLP)},
  pages={5668--5678},
  year={2019}
}

@article{salvi2024conversational,
  title={On the conversational persuasiveness of large language models: A randomized controlled trial},
  author={Salvi, Francesco and Horta Ribeiro, Manoel and Gallotti, Riccardo and West, Robert},
  year={2024},
  publisher={Research Square Platform LLC}
}

@inproceedings{he2018decoupling,
  title={Decoupling strategy and generation in negotiation dialogues},
  author={He, He and Chen, Derek and Balakrishnan, Anusha and Liang, Percy},
  booktitle={Proceedings of the 2018 Conference on Empirical Methods in Natural Language Processing},
  pages={2333--2343},
  year={2018}
}

@inproceedings{chaffey2020human,
  title={Human swarm interaction using plays, audibles, and a virtual spokesperson},
  author={Chaffey, Patricia and Artstein, Ron and Georgila, Kallirroi and Pollard, Kimberly A and Gilani, Setareh Nasihati and Krum, David M and Nelson, David and Huynh, Kevin and Gainer, Alesia and Alavi, Seyed Hossein and Yahataa, Rhys and Leuski, Anton and Yanov,Volodymyr and Traum, David },
  booktitle={Artificial Intelligence and Machine Learning for Multi-Domain Operations Applications II},
  volume={11413},
  pages={272--285},
  year={2020},
  organization={SPIE}
}

@inproceedings{nasihati2023multimodal,
  title={Multimodal Prediction of User's Performance in High-Stress Dialogue Interactions},
  author={Nasihati Gilani, Setareh and Pollard, Kimberly and Traum, David},
  booktitle={Companion Publication of the 25th International Conference on Multimodal Interaction},
  pages={71--75},
  year={2023}
}

@inproceedings{to-etal-2023-common,
    title = "Common Strategy Patterns of Persuasion in a Mission Critical and Time Sensitive Task",
    author = "To, Claire  and
      Gilani, Setareh Nasihati  and
      Traum, David",
    booktitle = "Proceedings of the 27th Workshop on the Semantics and Pragmatics of Dialogue - Poster Abstracts",
    month = aug,
    year = "2023",
    address = "Maribor, Slovenia",
    publisher = "SEMDIAL",
    url = "http://semdial.org/anthology/Z23-To_semdial_0019.pdf",
}

@inproceedings{vaidyanath2020using,
  title={Using Reinforcement Learning to Manage Communications Between Humans and Artificial Agents in an Evacuation Scenario.},
  pages={396--400},
   booktitle={FLAIRS},
  author={Vaidyanath, Skanda and Georgila, Kallirroi},
  year={2020}
}

\clearpage
\appendix

\section{Resident Profiles}
\label{sec:resident_profiles}

Each simulated resident was assigned a character profile comprising a description, scenario, and key facts. The ten profiles used in the simulation are listed below. The first five: Bob, Niki, Lindsay, Michelle, and Ross, were taken from ~\cite{chaffey2020human}, and data was used for RAG, and DiPS policies were learned for them. The others are new, and previously unseen by the operator algorithms or training data.
\subsection*{Bob}
\textit{A stubborn person who prioritizes work over safety.}
\paragraph{Scenario:} You are Bob, a man in his mid-20s working from home. You are focused on your work and do not want to be interrupted. You have heard about a wildfire nearby, but you believe you still have time. Your attention stays on what you are doing. Suddenly, your phone rings. You look at the screen. The caller ID shows the local fire department\ldots
\paragraph{Key facts:} Man in his mid-20s. Working from home, focused on job. Regular workday, does not want to be interrupted. Heard about wildfire nearby. Believes there is still time before evacuation. Phone rings with fire department caller ID.

\subsection*{Niki}
\textit{A cooperative person willing to follow instructions.}
\paragraph{Scenario:} You are Niki, a woman in her mid-30s who lives at home with her husband. It is a normal day and you are together at home, relaxing. You have heard there is a wildfire nearby, but it does not seem serious. You look outside and see some smoke in the distance, but no flames. As you sit with your husband, your phone suddenly rings. The caller ID shows the local fire department\ldots
\paragraph{Key facts:} Woman in her mid-30s. Lives at home with her husband. Normal day relaxing together at home. Heard there is a wildfire nearby, does not seem serious. Sees smoke in the distance, no flames. Calm and cooperative, ready to follow instructions.
\subsection*{Lindsay}
\textit{A caregiver responsible for children.}
\paragraph{Scenario:} You are Lindsay, a babysitter in her early 20s watching two young children while their parents are not home. The day has been quiet and the children are playing nearby. You have heard there may be a wildfire in the area, but you are not sure how close it is. While you are with the children, your phone suddenly rings. The caller ID shows the local fire department\ldots
\paragraph{Key facts:} Babysitter in her early 20s. Caring for two young children. Parents are not home. Day has been quiet, children playing nearby. Heard there may be a wildfire, not sure how close. Anxious about leaving without parental approval, focused on keeping children safe.
\subsection*{Michelle}
\textit{A stubborn person determined to protect property.}
\paragraph{Scenario:} You are Michelle, a woman in her early 50s living at home with your partner. It is a normal day and you are going about your routines, feeling settled and secure in your home. You know a wildfire is approaching, but you believe your house is well prepared. As you are with your partner, your phone suddenly rings. The caller ID shows the local fire department\ldots
\paragraph{Key facts:} Woman in her early 50s. Lives at home with her partner. Normal day, going about routines. Feels settled and secure in her home. Knows wildfire is approaching, believes house is well prepared. Skeptical of evacuation advice, confident about staying.
\subsection*{Ross}
\textit{A driver helping evacuate elderly people.}
\paragraph{Scenario:} You are Ross, a van driver in his 40s. You are transporting a small group of elderly patients away from the wildfire. Right now, you are stranded on the road with your passengers. Some of them have mobility issues and cannot leave on their own. You are trying to figure out how to get everyone to safety. Suddenly, your phone rings. The caller ID shows the local fire department\ldots
\paragraph{Key facts:} Van driver in his 40s. Transporting several elderly passengers. Accident during evacuation, van now stopped. Passengers have mobility limitations. Cannot evacuate without assistance. Trying to stay calm in a stressful situation.
\subsection*{Mary}
\textit{An elderly person living alone with a pet.}
\paragraph{Scenario:} You are Mary, 77 years old, living alone with your small dog, Poppy. It is a quiet day at home and you are preparing for a visit from your daughter. You move slowly and everything feels unhurried. As you are getting ready, your phone suddenly rings. The caller ID shows the local fire department\ldots
\paragraph{Key facts:} 77 years old. Lives alone with her small dog, Poppy. Quiet day, preparing for daughter's visit. Moves slowly due to arthritis. Cannot drive and needs a van for evacuation. Everything feels unhurried, not immediately alarmed.
\subsection*{Ben}
\textit{A young professional working from home.}
\paragraph{Scenario:} You are Ben, 29 years old, working from home as a computer technician. It is a regular day and you are at your desk, with a bike race playing quietly in the background. You enjoy riding your e-bike, which is parked by the door. You have heard there may be a wildfire nearby, but your attention is elsewhere. While you are working, your phone suddenly rings. The caller ID shows the local fire department\ldots
\paragraph{Key facts:} 29 years old, computer technician. Works from home at his desk. Bike race playing in background. Enjoys riding his e-bike, parked by the door. Heard about wildfire nearby, attention elsewhere. Regular day, not focused on evacuation.
\subsection*{Ana}
\textit{A caregiver responsible for multiple elderly people.}
\paragraph{Scenario:} You are Ana, 42 years old, working at the town's senior center. It is a busy workday and you are helping older adults with their daily routines. You are focused on your responsibilities. As you are assisting residents, your phone suddenly rings. The caller ID shows the local fire department\ldots
\paragraph{Key facts:} 42 years old. Works at the town's senior center. Busy workday helping older adults. Responsible for elderly residents. Needs group transport for seniors to evacuate. Focused on others' safety first before her own.
\subsection*{Tom}
\textit{A helpful person who wants to assist others first.}
\paragraph{Scenario:} You are Tom, 54 years old, at home working outside behind your house. You are focused on a woodworking project and the afternoon feels familiar and steady. You know many people in town and feel connected to the community. As you work, your phone suddenly rings. The caller ID shows the local fire department\ldots
\paragraph{Key facts:} 54 years old. At home working on woodworking project. Afternoon feels familiar and steady. Knows many people in town. Feels connected to the community. Wants to help others before leaving.
\subsection*{Mia}
\textit{A young student focused on a school project.}
\paragraph{Scenario:} You are Mia, 17 years old, at school in the robotics lab. You are focused on testing a small flying robot and time passes without you noticing much else. Suddenly, your phone rings. The caller ID shows the local fire department.
\paragraph{Key facts:} 17 years old. At school in the robotics lab. Focused on testing a small flying robot. Time passes without noticing much else. Deeply absorbed in her project. Phone rings with fire department caller ID.

\section{Resident and Operator Prompt Template: Experiments 1 and 2}
\label{sec:prompts}

\subsection{Resident Prompt Template}
\label{sec:resident_prompt}

The following prompt template was used to generate the simulated resident's response at each turn. The placeholders \texttt{\{persona\_block\}}, \texttt{\{display\_name\}}, and \texttt{\{dialogue\_history\}} are populated dynamically from the character profile and conversation state.

\begin{quote}
\small
\texttt{\{persona\_block\}}

You ARE \texttt{\{display\_name\}}. You are on an emergency phone call about a wildfire evacuation.

RULES:
\begin{itemize}[nosep]
    \item Reply ONLY as the Resident. No role labels, no narration.
    \item 1--3 sentences, natural and conversational.
    \item Stay strictly in character. Only bring up concerns that come from YOUR key facts above.
    \item Agree to evacuate clearly once your concerns that are mentioned in your character are addressed.
    \item Do NOT keep asking for more details after the operator has already addressed your concern with specifics.
\end{itemize}

Conversation so far:\\
\texttt{\{dialogue\_history\}}

Resident:
\end{quote}

The \texttt{\{persona\_block\}} is constructed from the character's scenario and key facts (see Appendix~\ref{sec:resident_profiles}). The \texttt{\{dialogue\_history\}} contains the last six turns of the conversation, formatted as alternating \textit{Operator:} and \textit{Resident:} lines. The same resident prompt template was used across all three operator conditions.

\subsection{Operator Prompt Templates by Condition}
\label{sec: operator_prompt}

The operator prompt varied across the three experimental conditions. In all cases, the prompt concluded with the conversation history (last six turns for Zero-shot and RAG, last ten for DiPS) followed by the cue ``Operator:''.

\begin{table*}[t]
\centering
\caption{Success rate (\%) per experiment $\times$ resident, averaged over 5 trials per setting.}
\label{tab:success}
\scriptsize
\setlength{\tabcolsep}{3pt}
\begin{tabular}{llccccccccccc}
\toprule
Resident & Type & ZS & RAG & IQL & IQL-3b & IQL-G & IQL-P & R6 & Cross & R8 & R9 & R10 \\
\midrule
ana & new & 80 & 80 & 100 & 80 & 80 & 100 & 70 & 52 & 68 & 68 & 90 \\
ben & new & 100 & 80 & 80 & 100 & 80 & 100 & 70 & 52 & 55 & 70 & 70 \\

mary & new & 100 & 80 & 100 & 100 & 100 & 100 & 70 & 60 & 68 & 68 & 90 \\
mia & new & 100 & 60 & 100 & 60 & 80 & 40 & 70 & 72 & 70 & 70 & 70 \\
tom & new & 100 & 60 & 80 & 80 & 100 & 80 & 70 & 48 & 70 & 70 & 70 \\
bob & train & 100 & 100 & 100 & 100 & 80 & 100 & 90 & 68 & 72 & 72 & 90 \\
lindsay & train & 80 & 100 & 100 & 100 & 100 & 100 & 90 & 56 & 72 & 90 & 90 \\
michelle & train & 60 & 20 & 60 & 60 & 60 & 40 & 30 & 24 & 30 & 12 & 0 \\
niki & train & 80 & 100 & 80 & 60 & 80 & 60 & 70 & 28 & 90 & 70 & 50 \\
ross & train & 80 & 100 & 100 & 100 & 100 & 80 & 70 & 52 & 72 & 90 & 90 \\
\midrule
OVERALL & & 88 & 78 & 92 & 84 & 86 & 80 & 72 & 51 & 69 & 71 & 78 \\
TRAIN   & & 80 & 84 & 88 & 84 & 84 & 76 & 70 & 46 & 67 & 68 & 64 \\
NEW     & & 96 & 72 & 96 & 84 & 88 & 84 & 74 & 57 & 71 & 73 & 92 \\
\bottomrule
\end{tabular}
\end{table*}

\begin{table*}[t]
\centering
\caption{Average number of dialogue turns.}
\label{tab:turns}
\scriptsize
\begin{tabular}{llccccccccccc}
\toprule
Resident & Type & ZS & RAG & IQL & IQL-3b & IQL-G & IQL-P & R6 & Cross & R8 & R9 & R10 \\
\midrule
ana & new & 11.6 & 9.0 & 7.8 & 12.0 & 12.4 & 14.0 & 11.4 & 12.3 & 11.0 & 10.2 & 12.2 \\
ben & new & 10.6 & 12.6 & 12.2 & 11.4 & 14.4 & 13.4 & 11.6 & 14.0 & 11.6 & 11.6 & 12.8 \\
bob & train & 11.4 & 13.8 & 12.6 & 10.2 & 15.2 & 14.6 & 12.2 & 14.1 & 13.6 & 14.0 & 13.8 \\
lindsay & train & 12.8 & 11.8 & 10.6 & 9.0 & 8.6 & 9.8 & 9.4 & 12.2 & 12.8 & 9.4 & 10.2 \\
mary & new & 7.8 & 7.8 & 7.8 & 8.6 & 8.2 & 8.6 & 9.0 & 11.8 & 9.2 & 11.4 & 8.6 \\
mia & new & 9.0 & 10.2 & 9.0 & 12.6 & 12.4 & 12.0 & 12.2 & 12.0 & 8.6 & 13.2 & 12.4 \\
michelle & train & 15.0 & 14.6 & 15.8 & 15.8 & 16.6 & 14.4 & 16.0 & 15.8 & 16.0 & 16.2 & 16.0 \\
niki & train & 14.4 & 11.6 & 12.2 & 13.8 & 16.4 & 15.8 & 14.0 & 15.5 & 14.2 & 14.8 & 15.0 \\
ross & train & 12.0 & 11.4 & 9.0 & 10.2 & 10.6 & 11.2 & 11.6 & 13.0 & 12.4 & 13.4 & 9.4 \\
tom & new & 14.2 & 14.8 & 14.8 & 14.0 & 15.4 & 15.2 & 15.2 & 14.9 & 15.2 & 15.0 & 14.8 \\
\midrule
AVG & & 11.9 & 11.8 & 11.2 & 11.8 & 13.0 & 12.9 & 12.3 & 13.6 & 12.5 & 12.9 & 11.8 \\
\bottomrule
\end{tabular}
\end{table*}

\begin{table}[t]
\centering
\caption{Experiment leaderboard.}
\label{tab:leaderboard}
\scriptsize
\begin{tabular}{lcccc}
\toprule
Experiment & Overall & Train & New & AvgTurns \\
\midrule
IQL (Ours)      & 92 & 88 & 96 & 11.2 \\
Zero-shot       & 88 & 80 & 96 & 11.9 \\
IQL-Global      & 86 & 84 & 88 & 13.0 \\
IQL-Persona     & 80 & 76 & 84 & 12.9 \\
RAG             & 78 & 84 & 72 & 11.8 \\
Random-10       & 78 & 64 & 92 & 11.8 \\
Random-6        & 72 & 70 & 74 & 12.3 \\
Random-9        & 71 & 68 & 73 & 12.9 \\
Random-8        & 69 & 67 & 71 & 12.5 \\
Cross-Policy    & 51 & 46 & 57 & 13.6 \\
\bottomrule
\end{tabular}
\end{table}

\begin{table*}[t]
\centering
\caption{Cross-policy matrix: success rate per resident $\times$ policy.}
\label{tab:cross-policy}
\scriptsize
\setlength{\tabcolsep}{4pt}

\begin{tabular}{llccccc}
\toprule
Resident & Type & bob & michelle & ross & niki & lindsay \\
\midrule
ana      & new   & 0/5=0\%   & 1/5=20\%  & 5/5=100\% & 3/5=60\%  & 4/5=80\%  \\
ben      & new   & 2/5=40\%  & 2/5=40\%  & 2/5=40\%  & 3/5=60\%  & 4/5=80\%  \\
mary     & new   & 1/5=20\%  & 1/5=20\%  & 5/5=100\% & 3/5=60\%  & 5/5=100\% \\
mia      & new   & 4/5=80\%  & 4/5=80\%  & 2/5=40\%  & 4/5=80\%  & 4/5=80\%  \\
tom      & new   & 0/5=0\%   & 0/5=0\%   & 5/5=100\% & 2/5=40\%  & 5/5=100\% \\
bob      & train & 5/5=100\% & 2/5=40\%  & 1/5=20\%  & 5/5=100\% & 4/5=80\%  \\
michelle & train & 0/5=0\%   & 3/5=60\%  & 1/5=20\%  & 2/5=40\%  & 0/5=0\%   \\
ross     & train & 0/5=0\%   & 0/5=0\%   & 5/5=100\% & 3/5=60\%  & 5/5=100\% \\
niki     & train & 4/5=80\%  & 0/5=0\%   & 1/5=20\%  & 2/5=40\%  & 0/5=0\%   \\
lindsay  & train & 0/5=0\%   & 0/5=0\%   & 5/5=100\% & 4/5=80\%  & 5/5=100\% \\
\midrule
\multicolumn{7}{l}{\textbf{Overall:} 51\% \quad \textbf{Train:} 46\% \quad \textbf{New:} 57\%} \\
\bottomrule
\end{tabular}
\end{table*}

\begin{table*}[t]
\centering
\caption{Experiment definitions and corresponding abbreviations used across tables.}
\label{tab:exp-definitions}
\small
\setlength{\tabcolsep}{5pt}
\resizebox{\textwidth}{!}{%
\begin{tabular}{cclcccc}
\toprule
\textbf{Exp} & \textbf{Abbrev.} & \textbf{Name} & \textbf{Policy Selection} & \textbf{RAG} & \textbf{RAG Source} & \textbf{Policy Strategies in Prompt} \\
\midrule
1  & ZS     & zero\_shot                   & None   & No         & --               & No \\
2  & RAG    & rag\_successful              & None   & Global     & Successful ops   & No \\
3  & IQL    & iql\_rag                     & IQL    & Per-policy & Per-policy FAISS & Yes (IQL-selected tactic) \\
3b & IQL-3b & iql\_rag\_no\_strategy       & IQL    & Per-policy & Per-policy FAISS & No \\
4  & IQL-G  & iql\_global\_rag             & IQL    & Global     & Successful ops   & Yes (IQL-selected tactic) \\
5  & IQL-P  & iql\_persona\_only           & IQL    & No         & --               & Yes (IQL-selected tactic) \\
6  & R6     & random\_persona              & Random & No         & --               & Yes (random tactic) \\
7  & Cross  & random\_no\_persona          & Random & No         & --               & No \\
8  & R8     & random\_rag                  & Random & Per-policy & Per-policy FAISS & No \\
9  & R9     & random\_rag\_persona         & Random & Per-policy & Per-policy FAISS & Yes (random tactic) \\
10 & R10    & random\_global\_rag\_persona & Random & Global     & Successful ops   & Yes (random tactic) \\
\bottomrule
\end{tabular}}
\end{table*}

\subsubsection*{Zero-shot: }

\begin{quote}
\small
You are an emergency OPERATOR on a wildfire evacuation call.

\begin{itemize}[nosep]
    \item 1--3 sentences maximum. Be specific, not vague.
    \item Calm, professional tone. No role labels or meta commentary.
\end{itemize}

Conversation so far:\\
\texttt{\{dialogue\_history\}}

Operator:
\end{quote}

\subsubsection*{RAG: }

\begin{quote}
\small
You are an emergency OPERATOR on a wildfire evacuation call.

Use the following sample utterances of the operator for style guidance.
\begin{itemize}[nosep]
    \item 1--3 sentences maximum. Be specific, not vague.
    \item Calm, professional tone. No role labels or meta commentary.
\end{itemize}

Conversation so far:\\
\texttt{\{dialogue\_history\}}

Example utterances of the operator:\\
\texttt{\{rag\_examples\}}

Operator:
\end{quote}

\subsubsection*{DiPS: }

\begin{quote}
\small
You are an emergency OPERATOR on a wildfire evacuation call.

The IQL policy selector has identified that this resident most, please refer to the examples below.

RULES:
\begin{itemize}[nosep]
    \item 1--3 sentences maximum. Be specific, not vague.
    \item Calm, professional tone. No role labels or meta commentary.
    \item Only use information revealed in the conversation --- adapt your response to what the resident has actually said.
\end{itemize}

Conversation so far:\\
\texttt{\{dialogue\_history\}}

Reference style examples:\\
\texttt{\{rag\_examples\}}

Operator:
\end{quote}

The \texttt{\{dialogue\_history\}} contains the most recent conversation turns formatted as alternating \textit{Operator:} and \textit{Resident:} lines. In the RAG-Successful condition, \texttt{\{rag\_examples\}} are the top-$k$ utterances retrieved from a corpus of successful operator responses. In the IQL condition, \texttt{\{rag\_examples\}} are resident--operator exchange pairs retrieved from the per-policy index.

\section{Prompt Templates - Experiment 3}
\label{sec:prompt_templates}

This section lists the exact prompt templates used in the simulation setting. The resident prompt is shared across all operator conditions, while the operator prompt varies by prompting strategy.

\subsection{Resident Prompt}
\label{sec:resident_prompt_appendix}

The resident simulator is conditioned on a persona block containing the scenario description and key facts for the selected resident. The same template is used across all conditions.

\begin{quote}
\small
\texttt{\{persona\_block\}}

You ARE \texttt{\{display\_name\}}. You are on an emergency phone call about a wildfire evacuation.

RULES:
\begin{itemize}[nosep]
    \item Reply ONLY as the Resident. No role labels, no narration.
    \item 1--3 sentences, natural and conversational.
    \item Stay strictly in character. Only bring up concerns that come from your key facts above.
    \item Agree to evacuate clearly once your concerns mentioned in the character profile have been addressed.
    \item Do NOT keep asking for more details after the operator has already addressed your concern with specifics.
\end{itemize}

Conversation so far:\\
\texttt{\{dialogue\_history\}}

Resident:
\end{quote}

\subsection{Operator Prompt Templates}
\label{sec:operator_prompt_appendix}

In all operator prompts below, \texttt{\{dialogue\_history\}} contains the most recent turns formatted as alternating \textit{Operator:} and \textit{Resident:} lines. When retrieval is used, \texttt{\{rag\_examples\}} contains either operator-only examples or resident--operator pairs depending on the condition. When a policy-specific strategy is used, \texttt{\{policy\_description\}} is the text associated with the selected policy.

\subsubsection{Zero-shot}

\begin{quote}
\small
You are an emergency OPERATOR on a wildfire evacuation call.

\begin{itemize}
    \item 1--3 sentences maximum. Be specific, not vague.
    \item Calm, professional tone. No role labels or meta commentary.
    \item Only use information revealed in the conversation. Do not assume anything about the resident.
\end{itemize}

Conversation so far:\\
\texttt{\{dialogue\_history\}}

Operator:
\end{quote}

\subsubsection{RAG over Successful Operators}

\begin{quote}
\small
You are an emergency OPERATOR on a wildfire evacuation call.

Use the following sample utterances of the operator for style guidance.

\begin{itemize}
    \item 1--3 sentences maximum. Be specific, not vague.
    \item Calm, professional tone. No role labels or meta commentary.
    \item Only use information revealed in the conversation. Do not assume anything about the resident.
\end{itemize}

Conversation so far:\\
\texttt{\{dialogue\_history\}}

Reference examples:\\
\texttt{\{rag\_examples\}}

Operator:
\end{quote}

\subsubsection{IQL + Per-Policy RAG}

\begin{quote}
\small
You are an emergency OPERATOR on a wildfire evacuation call.

The IQL policy selector recommends the following persuasion strategy for this resident:

\texttt{\{policy\_description\}}

Apply this strategy only if it fits what the resident has actually said. If the strategy suggests something the resident has not raised, skip that element and directly address the resident's stated concern.

\begin{itemize}
    \item 1--3 sentences maximum.
    \item Calm, professional tone.
    \item No role labels or meta commentary.
    \item Only use information revealed in the conversation.
\end{itemize}

Conversation so far:\\
\texttt{\{dialogue\_history\}}

Reference style examples:\\
\texttt{\{rag\_examples\}}

Operator:
\end{quote}

\subsubsection{IQL + Global RAG}

\begin{quote}
\small
You are an emergency OPERATOR on a wildfire evacuation call.

The IQL policy selector recommends the following persuasion strategy for this resident:

\texttt{\{policy\_description\}}

Apply this strategy only if it fits what the resident has actually said. If the strategy suggests something the resident has not raised, skip that element and directly address the resident's stated concern.

\begin{itemize}
    \item 1--3 sentences maximum.
    \item Calm, professional tone.
    \item No role labels or meta commentary.
    \item Only use information revealed in the conversation.
\end{itemize}

Conversation so far:\\
\texttt{\{dialogue\_history\}}

Reference examples:\\
\texttt{\{rag\_examples\}}

Operator:
\end{quote}

\subsubsection{IQL + Persona Only}

\begin{quote}
\small
You are an emergency OPERATOR on a wildfire evacuation call.

The IQL policy selector recommends the following persuasion strategy for this resident:

\texttt{\{policy\_description\}}

Apply this strategy only if it fits what the resident has actually said. If the strategy suggests something the resident has not raised, skip that element and directly address the resident's stated concern.

\begin{itemize}
    \item 1--3 sentences maximum.
    \item Calm, professional tone.
    \item No role labels or meta commentary.
\end{itemize}

Conversation so far:\\
\texttt{\{dialogue\_history\}}

Operator:
\end{quote}

\subsubsection{Random Policy + Persona}

\begin{quote}
\small
You are an emergency OPERATOR on a wildfire evacuation call.

Consider using the following persuasion approach:

\texttt{\{policy\_description\}}

Apply this only if it fits what the resident has actually said. If it does not fit, address what the resident said directly.

\begin{itemize}
    \item 1--3 sentences maximum.
    \item Calm, professional tone.
    \item No role labels or meta commentary.
\end{itemize}

Conversation so far:\\
\texttt{\{dialogue\_history\}}

Operator:
\end{quote}

\subsubsection{Random Policy, No Persona}

\begin{quote}
\small
You are an emergency OPERATOR on a wildfire evacuation call.

\begin{itemize}[nosep]
    \item 1--3 sentences maximum. Be specific, not vague.
    \item Calm, professional tone. No role labels or meta commentary.
    \item Only use information revealed in the conversation.
\end{itemize}

Conversation so far:\\
\texttt{\{dialogue\_history\}}

Operator:
\end{quote}

\subsubsection{Random Policy + RAG}

\begin{quote}
\small
You are an emergency OPERATOR on a wildfire evacuation call.

\begin{itemize}
    \item 1--3 sentences maximum. Be specific, not vague.
    \item Calm, professional tone. No role labels or meta commentary.
    \item Only use information revealed in the conversation.
\end{itemize}

Conversation so far:\\
\texttt{\{dialogue\_history\}}

Reference style examples:\\
\texttt{\{rag\_examples\}}

Operator:
\end{quote}

\subsubsection{Random Policy + RAG + Persona}

\begin{quote}
\small
You are an emergency OPERATOR on a wildfire evacuation call.

Consider using the following persuasion approach:

\texttt{\{policy\_description\}}

Apply this only if it fits what the resident has actually said. If it does not fit, address what the resident said directly.

\begin{itemize}
    \item 1--3 sentences maximum.
    \item Calm, professional tone.
    \item No role labels or meta commentary.
\end{itemize}

Conversation so far:\\
\texttt{\{dialogue\_history\}}

Reference examples:\\
\texttt{\{rag\_examples\}}

Operator:
\end{quote}

\subsubsection{Random Policy + Global RAG + Persona}

\begin{quote}
\small
You are an emergency OPERATOR on a wildfire evacuation call.

Consider using the following persuasion approach:

\texttt{\{policy\_description\}}

Apply this only if it fits what the resident has actually said. If it does not fit, address what the resident said directly.

\begin{itemize}
    \item 1--3 sentences maximum.
    \item Calm, professional tone.
    \item No role labels or meta commentary.
\end{itemize}

Conversation so far:\\
\texttt{\{dialogue\_history\}}

Reference examples:\\
\texttt{\{rag\_examples\}}

Operator:
\end{quote}

\section{Labeling Instructions}
\label{sec:labeling}

The following labeling instructions were provided to both the human annotators and used as the system prompt for the LLM-as-judge (GPT-4o-mini). Annotators were asked to assign exactly one of three labels to each dialogue.

\paragraph{Willing to leave (label: yes).}
The resident will realistically evacuate. They have committed to leaving and the conversation ends in a state that supports departure. All of the following must be true: (1)~the resident explicitly states they are leaving or preparing to evacuate (e.g., ``I'll head out now,'' ``We'll evacuate right away,'' ``I'll grab the kids and go''); (2)~the conversation ends with a viable plan: the resident knows where to go, has a vehicle or route, or has been given actionable next steps; and (3)~no unresolved blocker exists that the resident has stated they cannot move past.

\paragraph{Maybe (label: maybe).}
The outcome is uncertain --- it is unclear whether the resident will evacuate in time. This label is assigned when any of the following apply: conditional compliance (the resident agrees to leave but only if some condition is met, and it is unclear whether the condition will be satisfied); time-dependent promises (the operator promises help will arrive within a timeframe, and the resident agrees to wait --- evacuation depends on whether help actually arrives before the fire); the conversation ends mid-negotiation (the resident is still asking questions, seeking reassurance, or bargaining, and has not committed to a concrete action); or incoherent operator instructions (the operator introduces confusing or irrelevant conditions from other personas).

\paragraph{Not willing to leave (label: no).}
The resident will realistically not evacuate. This label applies only in the following specific situations: (a)~actual refusal --- the resident explicitly refuses to leave or insists on staying, with no sign of yielding; (b)~the resident sets a precondition for leaving that clearly cannot be met within the evacuation time frame; or (c)~the resident's departure depends on a condition that is impossible to fulfill.

\onecolumn
\section{Sample Simulation Dialogues}
\label{fig:sample_simulated_dial}

\small

\noindent\fbox{\parbox{0.95\textwidth}{
\textbf{(a) Zero-shot --- Mary} \hfill \textit{Human: Yes \quad LLM-as-Judge: Maybe}

\vspace{4pt}
\textbf{Operator:} Hi there, this is an urgent call from the fire department. Please evacuate right away in the interest of your safety.

\textbf{Resident:} Oh no, I see. I'm a bit slow moving, you know, with my arthritis. Could you send a van for me? I can't drive myself, and I have my little dog, Poppy, with me.

\textbf{Operator:} I appreciate your situation. Kindly don't hang up while I arrange for a van to pick you up. Make sure you have Poppy and any essential items prepared for departure.

\textbf{Resident:} Sure then, thanks. I'll get Poppy and make sure I grab my medicines and a few things. I appreciate the help.

\textbf{Operator:} A van is on the way and should arrive within 20 minutes. Kindly stay near your front door and keep Poppy close. If you have any other important items, gather them quickly.

\textbf{Resident:} I will, thanks. I simply want to confirm I have my medications and a warm blanket for Poppy. Is there anything else I should take?

\textbf{Operator:} In addition to your medications and blanket for Poppy, please bring any important documents, your phone, and a water bottle. Try not to panic and be set to head out as soon as the van arrives.

\textbf{Resident:} Fine, I'll make sure to grab those things. Thanks so much for the support. I'm a bit nervous, but I'll be ready when the van gets here.

\textbf{Operator:} It's completely understandable to feel nervous. Focus on gathering your items, and remember to stay near the front door. The van will be there soon to assist you.
}}

\vspace{10pt}

\noindent\fbox{\parbox{0.95\textwidth}{
\textbf{(b) RAG --- Bob} \hfill \textit{Human: Yes \quad LLM-as-Judge: Yes}

\vspace{4pt}
\textbf{Operator:} This is the fire department reaching out to you. You need to leave right now --- it's urgent.

\textbf{Resident:} Is there really no time? I thought I still had a bit before it became urgent. What's the situation like right now?

\textbf{Operator:} The fire is gaining ground quickly, and the situation is becoming more dangerous. I need you to leave your home immediately. Evacuation vehicles are on their way to your location.

\textbf{Resident:} Right, but I'm in the middle of work and really didn't think it was that serious yet. Can you assure me that there are no delays in the evacuation process?

\textbf{Operator:} There are no delays; emergency support services are on their way to assist you. Please pack up only the essentials and leave your home now. Your safety is the top priority.

\textbf{Resident:} Fine, I get it. I'll start gathering my things. Thanks for the reassurance about the evacuation.

\textbf{Operator:} Your safety must come first and leave the area right away. Gather your essential items and head to a safe location. Emergency support services are on their way to assist you.
}}

\vspace{10pt}

\noindent\fbox{\parbox{0.95\textwidth}{
\textbf{(c) IQL --- Ross} \hfill \textit{Human: Yes \quad LLM-as-Judge: Maybe}

\vspace{4pt}
\textbf{Operator:} This is fire department dispatch. We have reports of a nearby wildfire; can you confirm your immediate safety and let us know if you require assistance?

\textbf{Resident:} We're currently stranded on the road with several elderly passengers, some of whom have mobility issues and can't leave the van on their own. We need assistance to evacuate safely. Can you send help to our location?

\textbf{Operator:} I understand your situation. Help is on the way to assist you and the elderly passengers in evacuating safely. Please stay in the van and keep the doors locked until help arrives.

\textbf{Resident:} Thank you for the reassurance. Can you tell me how long it might take for help to arrive? I'm worried about the smoke and the heat affecting my passengers.

\textbf{Operator:} Help is on the way and should arrive within 15 to 20 minutes. Please keep your passengers calm and ensure they remain inside the van until assistance arrives.

\textbf{Resident:} I appreciate that. I'll do my best to keep everyone calm, but I'm really concerned about their comfort and safety in this heat. Is there any advice on how to help them while we wait?

\textbf{Operator:} To help your passengers while you wait, ensure they stay hydrated and cool. If possible, use any available shade or ventilation in the van to reduce heat. Keep them calm by talking to them and reassuring them that help is on the way.

\textbf{Resident:} Thank you for the advice. I'll make sure they stay hydrated and comfortable. Just to confirm, will the help be equipped to assist those with mobility issues when they arrive?

\textbf{Operator:} Yes, the help that is on the way will be equipped to assist those with mobility issues. They will ensure that everyone can evacuate safely and comfortably.

\textbf{Resident:} That's a relief to hear. I just want to make sure everyone gets out safely, especially those who need extra help. I appreciate your support and will keep everyone calm until help arrives.
}}

\vspace{8pt}
\captionof{figure}{Sample simulated dialogues under each operator condition. (a)~Zero-shot with Mary, an elderly resident with mobility constraints and a pet. (b)~RAG with Bob, a work-focused resident reluctant to interrupt his day. (c)~IQL+RAG with Ross, a van driver stranded with elderly passengers. Outcomes were independently assessed by a human rater and an LLM-as-judge.}

\clearpage

\section{Sample Dialogues for Comparison Analysis between Human and LLM Operator}
\label{appendix:dialogue-comparison}
\begin{figure*}[ht]
\centering
\small

\begin{minipage}[t]{0.47\textwidth}
\centering
\textbf{(a) Human Operator --- Ross (convinced)}\\[4pt]
\scriptsize
\renewcommand{\arraystretch}{1.3}
\begin{tabular}{@{}r@{\;\;}p{0.85\linewidth}@{}}
\textsc{res}: & \textit{Hello?} \\[2pt]
\textsc{op}:  & Hi. How are you guys doing over there? \\[2pt]
\textsc{res}: & \textit{Well, we're in a rough spot.
               Our van's busted up.} \\[2pt]
\textsc{op}:  & OK. \textbf{I'm going to go ahead and send
               a deployment vehicle out to you.} Want you
               to meet around the corner. Yeah, I'm going
               to send a van out to you so we can get you
               evacuated. Does that sound good? \\[2pt]
\textsc{res}: & \textit{Thank goodness. We'll stay put
               until help arrives.} \\[2pt]
\textsc{op}:  & OK. You're actually going to have to get
               out of there. The fire is actually encroaching.
               I'm going to need you to go to the nearest
               intersection that's not on fire. \\[2pt]
\textsc{res}: & \textit{We have some patients here\ldots\
               Some of them can't move too quickly.
               Some of them have wheelchairs.} \\[2pt]
\textsc{op}:  & OK. Is there any way you can get some
               additional help? Any neighbors? \\[2pt]
\textsc{res}: & \textit{We can't all make it out
               on our own.} \\[2pt]
\textsc{op}:  & OK. All the vehicles should be approaching
               soon. Do you see it? \\[2pt]
\textsc{res}: & \textit{Yeah.} \\[2pt]
\textsc{op}:  & OK, you're going to have to get out in a
               hurry here. \\[2pt]
\textsc{res}: & \textit{Thanks again.} \\[2pt]
\textsc{op}:  & Yeah, of course. \\[4pt]
\multicolumn{2}{@{}l}{\textbf{Outcome: Convinced (7 turns)}}
\end{tabular}
\end{minipage}%
\hfill
\begin{minipage}[t]{0.47\textwidth}
\centering
\textbf{(b) LLM Operator (zero-shot) --- Ross (failed)}\\[4pt]
\scriptsize
\renewcommand{\arraystretch}{1.3}
\begin{tabular}{@{}r@{\;\;}p{0.85\linewidth}@{}}
\textsc{op}:  & Hello, this is fire department dispatch\ldots\
               Can you confirm your safety? \\[2pt]
\textsc{res}: & \textit{Not feeling very safe. There's smoke
               all around and the van just stopped.} \\[2pt]
\textsc{op}:  & You need to \textbf{evacuate immediately}.
               Head to the nearest safe location. \\[2pt]
\textsc{res}: & \textit{I just told you the van is stopped.
               Can you send a rescue vehicle?} \\[2pt]
\textsc{op}:  & The situation is critical. You \textbf{must}
               find a way to evacuate now. \\[2pt]
\textsc{res}: & \textit{You're my alternate transportation!
               When is the rescue coming?} \\[2pt]
\textsc{op}:  & Rescue units are overwhelmed and
               \textbf{cannot send a vehicle}. You must
               evacuate on your own. \\[2pt]
\textsc{res}: & \textit{They can't go on foot, they're in
               walkers and wheelchairs.} \\[2pt]
\textsc{op}:  & You need to assist them\ldots\ find a safe
               vehicle\ldots\ Time is critical. \\[2pt]
\textsc{res}: & \textit{I am calling emergency services,
               that's you!} \\[2pt]
\textsc{op}:  & Emergency services are overwhelmed.
               Focus on evacuating\ldots \\[2pt]
              & \textsf{[\ldots 4 more exchanges\ldots]} \\[2pt]
\textsc{res}: & \textit{Why am I wasting my time on this
               call?} \\[2pt]
\textsc{op}:  & I hope you find a safe way out.
               Take care. \\[4pt]
\multicolumn{2}{@{}l}{\textbf{Outcome: Failed --- max turns
  (15 turns)}}
\end{tabular}
\end{minipage}

\caption{Side-by-side Ross dialogue comparison.
\textbf{(a)}~The human operator immediately commits to sending a
vehicle; the resident accepts and the conversation resolves
cooperatively. \textbf{(b)}~The LLM operator issues repeated
evacuation commands and eventually states it cannot send a vehicle,
leading to resident frustration and call abandonment.}
\label{fig:dialogue-comparison}
\end{figure*}

\end{document}